\documentclass[11pt,a4paper]{article}
\usepackage[hyperref]{naaclhlt2018}
\usepackage{times}
\usepackage{latexsym}
\usepackage{amsmath}
\usepackage{amsthm}
\usepackage{url}
\usepackage{breakurl}
\usepackage{array}
\usepackage{caption} 
\usepackage{wrapfig}
\usepackage{multirow}
\usepackage{setspace}
\usepackage{colordvi}  
\usepackage{graphicx}
\usepackage{epsfig}
\usepackage{epstopdf}
\usepackage{relsize}
\usepackage{dsfont}
\usepackage{enumitem}

\DeclareCaptionFont{10pt}{\fontsize{10pt}{12pt}\selectfont}
\aclfinalcopy 

\newcommand{\com}[1]{}
\newcommand{\samsa}{\textsf{SAMSA}}

\title{Semantic Structural Evaluation for Text Simplification}
\author{Elior Sulem, Omri Abend, Ari Rappoport \\
 Department of Computer Science, The Hebrew University of Jerusalem\\
{\tt \{eliors|oabend|arir\}@cs.huji.ac.il}
 }

\date{}

\begin{document}
\maketitle
\begin{abstract}

Current measures for evaluating text simplification systems focus on evaluating lexical text aspects, neglecting its structural aspects. In this paper we propose the first measure to address structural aspects of text simplification, called \samsa. It leverages recent advances in semantic parsing to assess simplification quality by decomposing the input based on its semantic structure and comparing it to the output. \samsa\ provides a reference-less automatic evaluation procedure, avoiding the problems that reference-based methods face due to the vast space of valid simplifications for a given sentence. Our human evaluation experiments show both \samsa's substantial correlation with human judgments, as well as the deficiency of existing reference-based measures in evaluating structural simplification.\footnote{All data and code are available in \url{https://github.com/eliorsulem/SAMSA}.}

\end{abstract}

\section{Introduction}

Text simplification ({\it TS}) addresses the translation of an input sentence into one or more simpler sentences. It is
a useful preprocessing step for several NLP tasks, such as machine translation \citep{C96,M14} and relation extraction \citep{N16}, and has also been shown useful in the development of reading aids, e.g., for people with dyslexia \citep{R13} or non-native speakers \citep{S02}.

The task has attracted much attention in the past decade \citep{Z10,WL11,W12,SA14,NG14},
but has yet to converge on an evaluation protocol that yields comparable results across
different methods and strongly correlates with human judgments. This is
in part due to the difficulty to combine the effects of different simplification operations (e.g., deletion, splitting and substitution). 
\citet{Xu16} has recently made considerable progress towards that goal,
and proposed to tackle it both by using an improved reference-based measure, 
named SARI, and by increasing the number of references.
However, their research focused on lexical, rather than structural simplification,
which provides a complementary view of TS quality as this paper will show.

This paper focuses on the evaluation of the structural aspects of the task. We introduce the semantic
measure \samsa\ (Simplification Automatic evaluation Measure through Semantic Annotation), the first
structure-aware measure for TS in general, and the first to use semantic structure
in this context in particular. \samsa\ stipulates that an optimal split of the input is
one where each predicate-argument structure is assigned its own sentence, and measures
to what extent this assertion holds for the input-output pair 
in question, by using semantic structure.
\samsa\ focuses on the core semantic components of the sentence, and is tolerant towards
the deletion of other units.\footnote{We do not consider other structural operations, such as passive to active transformations \citep{C02}, that are currently not treated by corpus-based simplification systems.}

For example, \samsa\ will assign a high score to the output split ``John got home. John gave Mary a call.'' for the input sentence ``John got home and gave Mary a call.'', as it splits each of its predicate-argument structures to a different sentence. Splits that alter predicate-argument relations such as ``John got home and gave. Mary called.'' are penalized by \samsa.

\samsa's use of semantic structures for TS evaluation has several motivations.
First, it provides means to measure the extent to which the meaning
of the source is preserved in the output. Second, it provides means
for measuring whether the input sentence was split to semantic units of
the right granularity.
Third, defining a semantic measure that does not require references
avoids the difficulties incurred by their non-uniqueness,
and the difficulty in collecting high quality references,
as reported by \citet{Xu15} and by \citet{NG14} with respect to the Parallel Wikipedia Corpus \citep[PWKP;][]{Z10}.
\samsa\ is further motivated by its use of semantic annotation only on the source side,
which allows to evaluate multiple systems using same source-side annotation,
and avoids the need to parse system outputs, which can be garbled.

In this paper we use the UCCA scheme for defining semantic structure \cite{AR13}. 
UCCA has been shown to be preserved remarkably well across translations \citep{S15} and has also been
successfully used for machine translation evaluation \citep{B16} (Section \ref{sec:related_work}). 
We note, however, that \samsa\ can be adapted to work with any semantic scheme that captures predicate-argument relations, such as AMR \citep{Ba13} or Discourse Representation Structures \citep{K81}, as used by \citet{NG14}.

We experiment with \samsa\ both where semantic annotation is carried out manually, and where it is carried out by a parser. See Section \ref{sec:samsa}.
We conduct human rating experiments and compare the resulting system rankings with
those predicted by \samsa. We find that \samsa's rankings obtain high
correlations with human rankings, and compare favorably to existing reference-based measures for TS.
Moreover, our results show that existing measures, which mainly target lexical simplification,
are ill-suited to predict human judgments where structural simplification is involved.
Finally, we apply \samsa\ to the dataset of the QATS shared task
on simplification evaluation \citep{S16}.
We find that \samsa\ obtains comparative correlation with human judgments on the task, despite operating in a more restricted setting, as it does not use human ratings as training data and focuses only on structural aspects of simplicity.
Section \ref{sec:related_work} presents previous work. Section \ref{sec:ucca} discusses UCCA. Section \ref{sec:samsa} presents \samsa. Section \ref{sec:human_evaluation} details the collection of human judgments.
Our experimental setup for comparing our human and automatic rankings is given in Section \ref{sec:experiments}, and results are given in Section \ref{sec:correlation}, showing superior results for \samsa. A discussion on the results is presented in Section \ref{sec:discussion}. Section \ref{sec:qats} presents experiments with \samsa\ on the QATS evaluation benchmark.

\vspace{-0.1cm}
\section{Related Work} \label{sec:related_work}
\vspace{-0.1cm}
\paragraph{Evaluation Metrics for Text Simplification.}
As pointed out by \citet{Xu16}, many of the existing measures for TS evaluation do not generalize across systems, 
because they fail to capture the combined effects of the different simplification operations. 
The two main directions pursued are direct human judgments and automatic measures borrowed from machine translation (MT) evaluation.
Human judgments generally include grammaticality (or fluency), meaning preservation (or adequacy) and simplicity. 
Human evaluation is usually carried out with a small number of sentences (18 to 20), 
randomly selected from the test set \citep{W12,NG14,NG16}.

The most commonly used automatic measure for TS is BLEU \citep{P02}.
Using 20 source sentences from the PWKP test corpus with 5 simplified sentences for each of them,
\citet{W12} investigated the correlation of BLEU with human evaluation, reporting positive correlation for
simplicity, but no correlation for adequacy.
\citet{S14} explored the correlation with human judgments of six automatic metrics: cosine
similarity with a bag-of-words representation, METEOR \citep{DL11}, TERp \citep{S09}, TINE \citep{R11} and two sub-components of TINE: T-BLEU (a variant of BLEU which uses lower n-grams when no 4-grams are found) and SRL (based on semantic role labeling). Using 280 pairs of a source sentence and a simplified output with only structural modifications, they found
positive correlations for all the metrics except TERp with respect to meaning preservation and positive albeit lower correlations for METEOR, T-BLEU and TINE with respect to grammaticality. Human simplicity judgments were not considered in this experiment. In this paper we collect human judgments for grammaticality, meaning preservation and {\it structural} simplicity. To our knowledge, this is the first work to target structural simplicity evaluation, and it does so both through elicitation of human judgments and through the definition of \samsa.

\citet{Xu16} were the first to propose two evaluation measures tailored for simplification, focusing on lexical simplification. The first metric is FKBLEU, a combination of iBLEU \citep{SZ12}, originally proposed for evaluating paraphrase generation by comparing the output both to the reference and to the input, and of the Flesch-Kincaid Index (FK), a measure of the readability of the text \citep{K75}. The second one is SARI (System output Against References and against the Input sentence) which compares the n-grams of the system output with those of the input and the human references,
separately evaluating the quality of words that are added, deleted
and kept by the systems. They found that FKBLEU and even more so SARI 
correlate better with human simplicity judgments than BLEU. On the other hand, BLEU (with multiple references) outperforms the other metrics on the dimensions of grammaticality and meaning preservation.  

As the Parallel Wikipedia Corpus (PWKP), usually used in simplification research, has been shown to contain a large portion of problematic simplifications \citep{Xu15,H15}, \citet{Xu16} further proposed to use multiple references (instead of a single reference) in the evaluation measures. \samsa\ addresses this issue by directly comparing the input and the output of the simplification system, without requiring manually curated references.

\vspace{-0.25cm}
\paragraph{Structural Measures for Text-to-text Generation.}
Other than measuring the number of splits \citep{NG14,NG16}, which only assesses the frequency of this operation and not its quality, no structural measures were previously proposed for the evaluation of structural simplification. The need for such a measure is pressing, given  recent interest in structural simplification, e.g., in the Split and Rephrase task \citep{N17}, which focuses on sentence splitting.

In the task of sentence compression, which is similar to simplification in that they both involve deletion and paraphrasing, \citet{CL06} showed that a metric that uses syntactic dependencies better correlates with human evaluation than a metric based on surface sub-strings. 
\citet{T16} found that structure-aware metrics obtain higher correlation with human evaluation over bigram-based metrics, in particular with grammaticality judgments, but that they do not significantly outperform bigram-based metrics on any parameter. Both \citet{CL06} and \citet{T16} use reference-based metrics that use syntactic structure on both the output and the references. \samsa\ on the other hand
uses linguistic annotation only on the source side, with semantic structures instead of syntactic ones.

Semantic structures were used in MT evaluation, for example in the MEANT metric \citep{L12}, which compares the output and the reference sentences, both annotated using SRL (Semantic Role Labeling). \citet{L14} proposes the XMEANT variant, which compares the SRL structures of the source and output (without using references). As some frequent constructions like nominal argument structures are not addressed by the SRL annotation, \citet{B16} proposed HUME, a human evaluation metric based on UCCA, using the semantic annotation only on the source side when comparing it to the output. We differ from HUME in proposing an automatic metric, tackling monolingual text simplification, rather than MT.

The UCCA annotation has also been recently used for the evaluation of Grammatical Error Correction (GEC). The US{\sc im} metric \citep{CA18} measures the semantic faithfulness of the output to the source by comparing their respective UCCA graphs.

\vspace{-0.25cm}
\paragraph{Semantic Structures in Text Simplification.}
In most of the work investigating the structural operations involved in text simplification, both in rule-based systems \citep{SA14} and in statistical systems \citep{Z10,WL11}, the structures that were considered were syntactic. \citet{NG14,NG16} proposed to use semantic structures in the simplification model, in particular in order to avoid splits and deletions which are inconsistent with the semantic structures. \samsa\ identifies such incoherent splits, e.g., a split of a phrase describing a single event, and penalizes them.

\citet{GS13} presented two simplification systems based on event extraction. One of them, named Event-wise Simplification, transforms each factual event motion into a separate sentence. This approach fits with \samsa's stipulation, that an optimal structural simplification is one where each (UCCA-) event in the input sentence is assigned a separate output sentence. However, unlike in their model, \samsa\ stipulates that not only should multiple events evoked by a verb in the same sentence be avoided in a simplification, but  penalizes sentences containing multiple events evoked by a lexical item of any category. 
For example, the sentence ``John's unexpected kick towards the gate saved the game'' which has two events, one evoked by ``kick'' (a noun) and another by ``saving'' (a verb) can be converted to ``John kicked the ball towards the gate. It saved the game.''
\vspace{-0.1cm}
\section{UCCA's Semantic Structures}\label{sec:ucca}
\vspace{-0.1cm}
In this section we will briefly describe the UCCA scheme, focusing on the concepts of Scenes and Centers which are key in the definition of \samsa. 
UCCA \citep[Universal Cognitive Conceptual Annotation;][]{AR13} is a semantic annotation scheme based on typological \citep{D10A,D10B,D12} and cognitive \citep{L08} theories which aims to represent the main semantic phenomena in the text, abstracting away from syntactic detail. UCCA structures are directed acyclic graphs whose nodes (or units) correspond either to the leaves of the graph (including the words of the text) or to several elements jointly viewed as a single entity according to some semantic or cognitive consideration. Unlike AMR, UCCA semantic units are directly anchored in the text \citep{AR17,B16}, which allows easy inclusion of a word-to-word alignment in the metric model (Section \ref{sec:samsa}).
\vspace{-0.2cm}
\paragraph{UCCA Scenes.}
A Scene, which is the most basic notion of the foundational layer of UCCA considered here, describes a movement, an action or a state which persists in time. Every Scene contains one main relation, which can be either a Process or a State. The Scene may contain one or more Participants, which are interpreted in a broad sense, including locations and destinations. For example, the sentence ``He ran into the park'' has a single Scene whose Process is ``ran''. The two Participants  are ``He'' and ``into the park''.

Scenes can have several roles in the text. First, they can provide additional information about an established entity (Elaborator Scenes) as for example the Scene ``who entered the house'' in the sentence ``The man who entered the house is John''. They can also be one of the Participants of another Scene, for example, ``he will be late'' in the sentence: ``He said he will be late''. In the other cases, the Scenes are annotated as parallel Scenes (H) which can be linked by a Linker (L): ``When$_L$ [he will arrive at home]$_H$, [he will call them]$_H$''.
\vspace{-0.1cm}
\paragraph{Unit Centers.}
With regard to units which are not Scenes, the category Center denotes the semantic head of the unit. For example, ``dogs'' is the center of the expression ``big brown dogs''  and ``box'' is the center of ``in the box''. There could be more than one  Center in a non-Scene unit, for example in the case of coordination, where all conjuncts are Centers.

 \vspace{-0.1cm}
 \section{The \samsa\ Metric}\label{sec:samsa}

\samsa's main premise is that a structurally correct simplification is one where:
(1) each sentence contains a single event from the input (UCCA Scene), (2) the main relation of each of the events and
their participants are retained in the output.

For example, consider ``John wrote a book. I read that book.'' as a
simplification of ``I read the book that John wrote.''.
Each output sentence contains one Scene, which has the same Scene elements as the source, and would thus be deemed correct by \samsa.
On the other hand, 
the output ``John wrote. I read the book.'' is an incorrect split of that sentence, since a participant of the ``writing'' Scene, namely ``the book'' is absent in the split sentence. \samsa\ would indeed penalize such a case.

Similarly, Scenes which have elements across several sentences receive a zero score by \samsa. As an example, consider
the sentence ``The combination of new weapons and tactics marks this battle as the end of chivalry'', and erroneous
split ``The combination of new weapons and tactics. It is the end of chivalry.'' (adapted from the output of a recent system
on the PWKP corpus), which does not preserve the original meaning.

\vspace{-0.1cm}
\subsection{Matching Scenes to Sentences} \label{sc2s}

\samsa\ is based on two external linguistic resources.
One is a semantic annotation (UCCA in our experiments) of the source side, which can be
carried out either manually or automatically, using the TUPA parser\footnote{\url{https://github.com/danielhers/tupa}} 
\citep[Transition-based UCCA parser;][]{H17} for UCCA.
UCCA decomposes each sentence $s$ into a set of Scenes $\{sc_1,sc_2,..,sc_n\}$, where each scene $sc_i$ contains
a main relation $mr_i$ (sub-span of $sc_i$) and a set of zero or more participants $A_i$.

The second resource is a word-to-word alignment $A$ between the words in the input and one or zero words in the output. 
The monolingual alignment thus permits \samsa\ not to penalize outputs that involve
lexical substitutions (e.g., ``commence'' might be aligned with ``start''). We denote by $n_{inp}$ the number of UCCA
Scenes in the input sentence and by $n_{out}$ the number of sentences in the output. 

Given an input sentence's UCCA Scenes $sc_1,\ldots, sc_{n_{inp}}$, a non-annotated output of a
simplification system split into sentences $s_1,\ldots,s_{n_{out}}$,
and their word alignment $A$, we distinguish between two cases:
\vspace{-0.2cm}
\begin{enumerate}[wide, labelwidth=!, labelindent=0pt, leftmargin=10pt]

\item
  {$n_{inp} \ge n_{out}$}: 
  in this case, we compute the maximal Many-to-1 correspondence between Scenes and sentences.
  A Scene is matched to a sentence in the following way. We say that a leaf $l$ in a
  Scene $sc$ is {\it consistent} in a Scene-sentence mapping $M$ which maps $sc$ to
  a sentence $s$, if there is a word $w \in s$ which $l$ aligns to (according to the word alignment $A$).
  The score of matching a Scene $sc$ to a sentence $s$ is then defined to be the total number of
  consistent leaves in $sc$. We traverse the Scenes in their order of occurrence in the text, selecting for
  each the sentence that maximizes the score.
  If $n_{inp} = n_{out}$, once a sentence is matched to a Scene, it cannot be matched to another one. Ties between sentences
  are broken towards the sentence that appeared first in the output.

         \begin{small}
         $ \ \ \ \ \ \ \ \ \ \ \ \ \ {\rm M^*}(sc_i) = {\rm argmax}_{s} score(sc_i,s)$ \\
        ${\rm s.t.}\,\,\, s \notin \{M^*(sc_1),\dots,M^*(sc_{i-1})\}\ \ {\rm if} \  n_{inp} = n_{out}$
        \end{small}

\item {$n_{inp}<n_{out}$}: In this case, a Scene will necessarily be split across several sentences.
  As this is an undesired result, we assign this instance a score of zero. 

\end{enumerate}

\vspace{-0.25cm}
\subsection{Score Computation}

\paragraph{Minimal Centers.}

The minimal center of a UCCA unit $u$ is UCCA's notion of a semantic head word, 
defined through recursive rules not unlike the head propagation
rules used for converting constituency structures to dependency structures. 
More formally, we define the minimal center of a UCCA unit $u$ (here a Participant or a Main Relation) to be the UCCA graph's leaf reached by starting from $u$ and iteratively selecting the child tagged as Center.
If a Participant (or a Center inside a Participant) is a Scene, its center is the main relation
(Process or State) of the Scene.

For example, the center of the unit ``The previous president of the commission'' ($u_1$) is ``president of the commission''. The center of the latter is ``president'', which is a leaf in the graph. So the minimal center of $u_1$ is ``president''.

\paragraph{}
Given the input sentence Scenes $\{sc_1,...,sc_{n_{inp}}\}$, the output sentences
$\{s_1,...,s_{n_{out}}\}$, and a mapping between them $M^*$, \samsa\ is defined as:

\vspace{-.5cm} 

\begin{scriptsize}
\begin{equation*}
   \dfrac{n_{out}}{n_{inp}} \dfrac{1}{2n_{inp}} \sum_{sc_i} \big[\mathds{1}_{M^*(sc_i)}(MR_i) + \dfrac{1}{k_i} \sum_{j=1}^{k_i} \mathds{1}_{M^*(sc_i)}({\rm Par}^{(j)}_{i}) \big]
\end{equation*}
\end{scriptsize}

\vspace{-.3cm}

where ${\rm MR_i}$ is the minimal center of the main relation (Process or State) of $sc_i$,
and ${\rm Par^{(j)}_i}$ ($j=1,\dots,k_i$) are the minimal centers of the Participants of $sc_i$.

For an output sentence $s$, $\mathds{1}_s(u)$ is a function from the input units to $\{0,1\}$,
which returns 1 iff $u$ is aligned (according to $A$) with
a word in $s$.\footnote{In some cases, the unit $u$ can be a sequence of centers (if there are several minimal centers). In these cases, $\mathds{1}_s(u)$ returns 1 iff the condition holds for all centers.}

The role of the non-splitting penalty term $n_{out}/n_{inp}$ in the \samsa\ formula is to penalize cases where
the number of sentences in the output is smaller than the number of Scenes.
In order to isolate the effect of the non-splitting penalty, we experiment with an additional metric
\samsa$_{abl}$\ (reads ``\samsa\ ablated''), which is identical to \samsa\, but does not take this term into account.
Corpus-level \samsa\ and \samsa$_{abl}$\ scores are obtained by averaging their sentence scores.

In the case of implicit units i.e. omitted units that do not appear explicitly in the text \citep{AR13},
since the unit preservation cannot be directly captured, the score $t$ for the relevant unit will be set
to $0.5$. For example, in the Scene ``traveling is fun'', the people who are traveling correspond to
an implicit Participant.
As implicit units are not covered by TUPA, this will only be relevant for the semi-automatic implementation of the metric (see Section \ref{sec:experiments}).

\vspace{-0.1cm}
\section{Human Evaluation Benchmark} \label{sec:human_evaluation}

\subsection{Evaluation Protocol}
For testing the automatic metric, we first build a human evaluation benchmark, 
using 100 sentences from the complex part of the PWKP corpus and the outputs of
six recent simplification systems for these sentences:\footnote{All the data can be found here: \url{http://homepages.inf.ed.ac.uk/snaraya2/data/simplification-2016.tgz}.}
(1) TSM \citep{Z10} using Tree-Based SMT,
(2) RevILP \citep{WL11} using Quasi-Synchronous Grammars, (3) PBMT-R \citep{W12} using Phrase-Based SMT,
(4) Hybrid \citep{NG14}, a supervised system using DRS,
(5) UNSUP \citep{NG16}, an unsupervised system using DRS, and 
(6) Split-Deletion \citep{NG16}, the unsupervised system with only structural operations.

All these systems explicitly address at least one type of structural simplification operation.
The last system, Split-Deletion, performs only structural (i.e., no lexical) operations.
It is thus an interesting test case for \samsa\, since here the aligner can be replaced by
a simple match between identical words. In total we obtain 600 system outputs from the six systems, as well as
100 sentences from the simple Wikipedia side of the corpus, which serve as references.
Five in-house annotators with high proficiency in English evaluated the resulting 700 input-output pairs
by answering the questions in Table \ref{questions}.\footnote{Each input-output pair was rated by all five annotators.}

Qa addresses grammaticality, Qb and Qc capture two complementary aspects of meaning preservation
(the addition and the removal of information) and Qd addresses structural simplicity.
Possible answers are: 1 (``no''), 2 (``maybe'') and 3 (``yes'').
Following \citet{GS13}, we used a 3 point Likert scale, which has recently been shown
to be preferable over a 5 point scale through human studies on sentence compression \citep{T16}.

Question Qd was accompanied by a negative example
\footnote{Other questions appeared without any example.}
 showing a case of lexical simplification, where a complex word is replaced by a
simple one. A positive example was not included so as not to bias the annotators by
revealing the nature of the operations our experiments focus on (i.e., splitting and deletion).
 
The PWKP test corpus \citep{Z10} was selected for our experiments over the development and test sets used in \cite{Xu16},
as the latter's selection process was explicitly biased towards input-output pairs that mainly contain lexical
simplifications. 

\vspace{-0.1cm}
\begin{center}
\begin{table}[h]
\scriptsize
\begin{center}
\begin{tabular}{|l|l|}
\hline
Qa & \parbox{0.6\linewidth}{Is the output grammatical?}\\
\hline
Qb & \parbox{0.6\linewidth}{Does the output add information, compared to the input?}\\
\hline
Qc &\parbox{0.6\linewidth}{Does the output remove important information, compared to the input?}\\
\hline
Qd &\parbox{0.6\linewidth}{Is the output simpler than the input, ignoring the complexity of the words?}\\
\hline
\end{tabular}
\end{center}
\hfill
\caption{Questions for the human evaluation}
\label{questions} 
\end{table}
\end{center}

\vspace{-0.5cm}
\subsection{Human Score Computation}

Given the annotator's answers, we consider the following scores. First, the grammaticality score 
$\mathcal{G}$ is the answer to Qa. By inverting (changing 1 to 3 and 3 to 1) the answer for Qb, we obtain 
a Non-Addition score indicating to which extent no additional information has been added. Similarly, 
inverting the answer to Qc yields the Non-Removal score.
Averaging these two scores, we obtain the meaning preservation score $\mathcal{P}$. 
Finally, the structural simplicity score $\mathcal{S}$ is the answer to Qd. Each of these scores is averaged over
the five annotators. We further compute an average human score:

{\small $${\rm AvgHuman} = \frac{1}{3} (\mathcal{G} + \mathcal{P} + \mathcal{S})$$}

\vspace{-0.5cm}

\subsection{Inter-annotator Agreement}

Inter-annotator agreement rates are computed in two ways.
Table \ref{absolute_agreement} presents the absolute agreement and
Cohen's quadratic weighted $\kappa$ \citep{C68}. 
Table \ref{spearman_agreement} presents Spearman's correlation ($\rho$) between the human ratings of the input-output pairs (top row), and between the resulting system scores (bottom row).
In both cases, the agreement between the five annotators is computed
as the average agreement over the 10 annotator pairs.

\vspace{-0.1cm}
\begin{center}
\begin{table}[ht]
\scriptsize
\centering
\setlength\tabcolsep{5pt}
\begin{tabular}{|c|c|c|c|c|}
\hline
& {\bf Qa} & {\bf Qb}& {\bf Qc} & {\bf Qd} \\
\hline\hline
{\bf Total} &0.58 (0.56)   &0.74 (0.30)& 0.53 (0.45) &0.57 (0.10)\\
\hline\hline
{\bf TSM} & 0.59 (0.47) &0.75 (0.27) &0.50 (0.40) &0.43 (0.08)\\
\hline
{\bf RevILP} &0.61 (0.59)  & 0.78 (0.27)& 0.60 (0.43) &0.62 (0.11)\\
\hline 
{\bf PBMT-R} &0.47 (0.42)  &0.70 (0.20) &0.58 (0.31) & 0.76 (0.10)\\
\hline
{\bf Hybrid} &0.59 (0.46) & 0.77 (0.26) & 0.52 (0.48) & 0.72 (0.15)\\
\hline
{\bf UNSUP} &0.51 (0.42) &0.59 (0.10) &0.45 (0.17) &0.52 (0.04)\\
\hline 
{\bf Split-Deletion}  &0.59 (0.48) & 0.93 (0.02) &0.45 (0.29) & 0.55 (0.04)\\
\hline
    {\bf Reference} &0.70 (0.40) & 0.66 (0.46) & 0.52 (0.58) & 0.41 (0.12)\\
\hline
\end{tabular}
\hfill
\caption{Inter-annotator absolute agreement (and quadratic weighted $\kappa$), averaged over the 10 annotator pairs. Rows correspond to systems, columns to questions. The top ``Total'' row refers to the concatenation of the outputs of all 6 systems together with the reference sentences. 
\label{absolute_agreement}}
\end{table}
\end{center}

\vspace{-1cm}

\begin{center}
\begin{table}[ht]
\scriptsize
\centering
\begin{center}
\setlength\tabcolsep{5pt}  
\begin{tabular}{|c|c|c|c|c|c|}
\hline
& {\bf Qa} & {\bf Qb}& {\bf Qc} & {\bf Qd} & {\rm AvgHuman} \\
\hline
{\bf Sen.} &0.63$^*$ &0.30$^*$ & 0.48$^*$ & 0.11$^{**}$ & 0.49$^*$ \\
\hline
{\bf Sys.} & 0.92$^{**}$& 0.54 (0.1)& 0.64 (0.06)& 0.14 (0.4)& 0.64 (0.06)\\
\hline 
\end{tabular}
\end{center}
\hfill
\caption{Spearman's correlation (and $p$-values) of the system-level (top row) and sentence-level (bottom row) ratings of the five annotators. $^* p < 10^{-5}$, $^{**}p=0.002$.}
\label{spearman_agreement}
\end{table}
\end{center}
\vspace{-0.6cm}
\section{Experimental Setup}\label{sec:experiments}

We further compute \samsa\ for the 100 sentences of the PWKP test set and the corresponding system outputs.
Experiments are conducted in two settings:
(1) a semi-automatic setting where UCCA annotation was carried out
manually by a single expert UCCA annotator using the UCCAApp annotation software \cite{A17},
and according to the standard annotation guidelines;\footnote{\url{http://www.cs.huji.ac.il/~oabend/ucca.html}}
(2) an automatic setting where the UCCA annotation was carried out by
the TUPA parser \citep{H17}.
Sentence segmentation of the outputs was carried out using the
NLTK package \citep{LB02}. 
For word alignments, we used the aligner of \citet{Sultan14}.\footnote{\url{https://github.com/ma-sultan/monolingual-word-aligner}}

\begin{center}
\begin{table*}[ht]
\scriptsize
\centering
\begin{tabular}{|c|c|c|c|c||c|c|c||c|c|}
  \hline
 & \multicolumn{4}{c||}{\bf Reference-less} & \multicolumn{3}{c||}{\bf Reference-based } & \multicolumn{2}{c|}{\bf $\Delta$ from Source} \\
  \hline
 & \multicolumn{2}{c|}{\bf \samsa} & \multicolumn{2}{c||}{\bf \samsa$_{abl}$} & {\bf BLEU} & {\bf SARI} & {\bf -LD$_{{\rm SR}}$} & {\bf -LD$_{{\rm SC}}$} & {\bf \# Split Sents.}\\
&  Semi-Auto. & Auto. &Semi-Auto. &Auto. & & & & & \\
\hline
{\bf $\mathcal{G}$} &{\bf 0.54} (0.1) &   0.37 (0.2) &  0.14 (0.4) &  0.14 (0.4) & 0.09 (0.4) & -0.77 (0.04) & -0.43 (0.2)  & -0.09 (0.4) & 0.09 (0.4)\\
\hline     
{\bf $\mathcal{P}$} &-0.09 (0.4) & -0.37 (0.2) &  {\bf 0.54} (0.1) & {\bf 0.54} (0.1) & 0.37 (0.2) & -0.14 (0.4) & 0.03 (0.5) & 0.37 (0.2)& -0.49 (0.2)\\
\hline
{\bf $\mathcal{S}$} &  0.54 (0.1) &   0.71 (0.06)& -0.71 (0.06) & -0.71 (0.06)  & -0.60 (0.1) & -0.43 (0.2) &  -0.43 (0.2) & -0.54 (0.1) & {\bf 0.83} (0.02)\\
\hline
{\bf AvgHuman} & {\bf 0.58} (0.1) &  0.35 (0.1) & 0.09 (0.2) & 0.09 (0.2) & 0.06 (0.5) & -0.81 (0.02) &  -0.46 (0.2)& -0.12 (0.4)&  0.14 (0.4)\\
\hline
\end{tabular}
\hfill
\caption{\label{tab:spearman_system}
  Spearman's correlation of system scores i.e. Pearson's correlation of system rankings (and $p$-values), between evaluation measures (columns) and human judgments (rows). The ranking is between the six simplification systems experimented with. The left block of columns corresponds to the \samsa\ and \samsa$_{abl}$ measures, in their semi-automatic and automatic forms. The middle block of columns corresponds to the reference-based measures SARI and BLEU, as well as -LD$_{{\rm SR}}$, which is the negative Levenshtein distances of the system output from the reference. The right block corresponds to measures of conservatism, and reflect how well the tendency of the systems to introduce changes to the input correlates with the human rankings. The block includes -LD$_{{\rm SC}}$, the negative Levenshtein distance from the source sentence, and the number of input sentences split by each of the systems.
  Levenshtein distances are taken as negative in order to capture similarity between the output and source/reference.
  The measure with the highest correlation in each row is boldfaced.
}

\end{table*}
\end{center}

\vspace{-0.8cm}
\section{Correlation with Human Evaluation} \label{sec:correlation}

We compare the system rankings obtained by \samsa\ and by the four human parameters. We find
that the two leading systems according to {\rm AvgHuman} and \samsa\ turn out to be the same:
Split-Deletion and RevILP. This is the case both for the semi-automatic and the
automatic implementations of the metric.
A Spearman $\rho$ correlation between the human and \samsa\ scores (comparing their rankings)  is presented in
Table \ref{tab:spearman_system}.

We compare \samsa\ and \samsa$_{abl}$\ to the reference-based measures SARI\footnote{Data and code for can be found in \url{https://github.com/cocoxu/simplification}.} \cite{Xu16} and BLEU, as well as to
the negative Levenshtein distance to the reference (-LD$_{{\rm SR}}$). We use the only
available reference for this corpus, in accordance with the standard practice.
SARI is a reference-based measure, based on n-gram overlap between the source, output and reference,
and focuses on lexical (rather than structural) simplification.
For completeness, we include the other two measures reported in \citet{NG16}, which are measures of
similarity to the input (i.e., they quantify the tendency of the systems to introduce changes to the
input): the negative Levenshtein distances between the output and input
compared to the original complex corpus (-LD$_{{\rm SC}}$), and
the number of sentences split by each of the systems.

The highest correlation with {\rm AvgHuman} and grammaticality
is obtained by semi-automatic \samsa\ (0.58 and 0.54), a high correlation
especially in comparison to the inter-annotator agreement
on {\rm AvgHuman}  (0.64, Table \ref{spearman_agreement}).
The automatic version obtains high correlation with human judgments in these settings,
where for structural simplicity, it scores somewhat higher than the semi-automatic
\samsa.
The highest correlation with structural simplicity is obtained by the number of sentences with splitting,
where \samsa\ (automatic and semi-automatic) is second and third highest,
although when restricted to multi-Scene sentences, the correlation for \samsa\
(semi-automatic) is higher (0.89, $p=0.009$ and 0.77, $p=0.04$).

The highest correlation for meaning preservation is obtained by \samsa$_{abl}$\, which provides
further evidence that the retainment of semantic structures is a strong predictor of
meaning preservation \cite{S15}.
\samsa\ in itself does not correlate with meaning preservation, probably due to its
penalization of under-splitting sentences. 

Note that the standard reference-based measures for simplification, BLEU and SARI, obtain
low and often negative correlation with human ratings.
We believe that this is the case because SARI and BLEU admittedly focus on lexical simplification,
and are difficult to use to rank systems which also perform structural simplification.

Our results thus suggest that \samsa\ provides additional value
in predicting the quality of a simplification
system and should be reported in tandem with more lexically-oriented measures.

\vspace{-0.1cm}
\section{Discussion} \label{sec:discussion}
\vspace{-0.1cm}
\paragraph{Human evaluation parameters.}
The fact that the highest correlations for structural simplicity and meaning preservation are obtained by different metrics (\samsa\ and \samsa$_{abl}$\ respectively)
highlights the complementarity of these two parameters for evaluating TS quality but also the difficulty of capturing them together. Indeed, a given sentence-level operation could both change the original meaning by adding or removing information (affecting the $\mathcal{P}$ score) and increase simplicity ($\mathcal{S}$). On the other hand, the identity transformation perfectly preserves the meaning of the original sentence without making it simpler.

For examining this phenomenon, we compute Spearman's correlation at the system-level between the simplicity and meaning preservation human scores. We obtain a correlation of -0.77 ($p = 0.04$) between $\mathcal{S}$ and $\mathcal{P}$. The correlation between $\mathcal{S}$ and the two sub-components of $\mathcal{P}$, the Non-Addition and the Non-Removal scores, are -0.43 ($p = 0.2$) and -0.77 ($p = 0.04$) respectively. These negative correlations support our use of an average human score for assessing the overall quality of the simplification. 

\begin{figure*}[t!]
\centering
\includegraphics[width=0.49\textwidth]{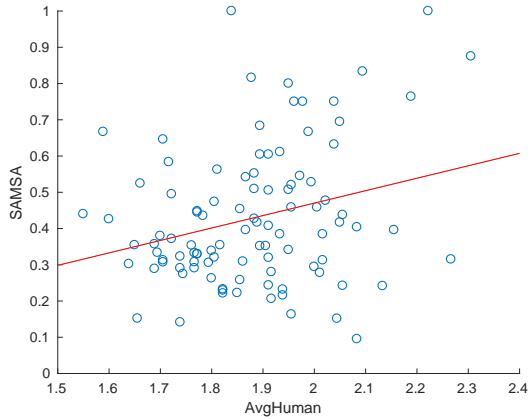}
\caption{Joint distribution of the automatic \samsa\ and the ${\rm AvgHuman}$ scores at the sentence level. Each point in the graph corresponds to a single source sentence. In addition to the scatter plot, a least-squares regression line is presented.}
\label{figure:joint_distribution}
\end{figure*}

\paragraph{Distribution at the sentence level.}
In addition to the system-level analysis presented in Section \ref{sec:correlation}, we also investigate the behavior of \samsa\ at the sentence level by examining its joint distribution with the human evaluation scores. Focusing on the ${\rm AvgHuman}$ score and the automatic implementation of \samsa\ and using the same data as in Section \ref{sec:correlation}, we consider a single pair of scores $({\rm AvgHuman}_{i}, \samsa_{i}), \ 1 \leq i \leq 100$,  for each of the 100 source sentences, averaging over the \samsa\ and human scores obtained for the 6 simplification systems (See Figure \ref{figure:joint_distribution}).

The joint distribution indicates a positive correlation between \samsa\ and ${\rm AvgHuman}$. The corresponding Pearson correlation is indeed 0.27 ($p=0.03$).
 
\vspace{-0.1cm}
\section{Evaluation on the QATS Benchmark} \label{sec:qats}
\vspace{-0.1cm}
In order to provide further validation for \samsa\ predictive value for quality of simplification systems, we report
\samsa's correlation with a recently proposed benchmark, used for the QATS (Quality Assessment for Text Simplification) 
shared task \citep{S16}.
\vspace{-0.2cm}

\paragraph{Setup.} The test corpus contains 126 sentences taken from 3 datasets described in \citet{S16}\footnote{\url{http://qats2016.github.io/shared.html}}:
(1) EventS: original sentences from the EMM News-Brief\footnote{\url{emm.newsbrief.eu/NewsBrief/clusteredition/en/latest.html}} and their syntactically simplified versions (with significant content reduction) 
by the EventSimplify TS system \citep{GS13}\footnote{\url{takelab.fer.hr/data/symplify}} (the test corpus contains 54 pairs from this dataset),
(2) EncBrit: original sentences from the Encyclopedia Britannica \citep{BE03} and their automatic simplifications obtained using ATS systems based on several phrase-based statistical MT systems \citep{Sa15} trained on Wikipedia TS corpus \citep{CK11} (24 pairs), and 
(3) LSLight: sentences from English Wikipedia and their automatic simplifications \citep{GS15} 
by three different lexical simplification systems \citep{B11,H14,GS15} (48 pairs).

Human evaluation is also provided by this resource, with scores for overall quality, grammaticality, meaning preservation
and simplicity. Importantly, the simplicity score does not explicitly refer to the output's structural simplicity, but rather
to its readability. We focus on the overall human score, and compare it to \samsa.
Since different systems were used to simplify different portions of the input, correlation is taken at the sentence level.

We use the same implementations of \samsa. Manual UCCA annotation is here performed by one of the authors of this paper.

\vspace{-0.25cm}
\paragraph{Results.}
We follow \citet{S16} and report the Pearson correlations (at the sentence level) between the rankings of the metrics and the human evaluation scores. Results show that the semi-automatic/automatic \samsa\ obtains a Pearson correlation of 0.32 and 0.28 with the human scores. This places these measures in the 3rd and 4th places in the shared task, where the only two systems that surpassed it are
marginally better, with scores of 0.33 and 0.34, and where the next system in QATS obtained a correlation of 
0.23.

This correlation by \samsa\ was obtained in more restricted conditions, compared to the measures that competed in QATS.
First, \samsa\ computes its score by only considering the UCCA structure of the source, and an automatic word-to-word alignment between the source and output. Most QATS systems, including OSVCML and OSVCML2 \cite{NS16} which scored highest on the shared task, use an ensemble
of classifiers based on bag-of-words, POS tags, sentiment information, negation, readability measures and other resources. Second, the
systems participating in the shared task had training data available to them, annotated by the same annotators as the test data. 
This was used to train classifiers for predicting their score.
This gives the QATS measures much predictive strength, but hampers their interpretability. 
\samsa\ on the other hand is conceptually simple and interpretable.
Third, the QATS shared task does not focus on structural simplification, but experiments on different types of systems.
Indeed, some of the data was annotated by systems that exclusively perform lexical simplification, 
which is orthogonal to \samsa's  structural focus. 

\vspace{-0.2cm}
\paragraph{}
Given these factors, \samsa's competitive correlation with the participating systems in QATS suggests that
structural simplicity, as reflected by the correct splitting of UCCA Scenes, captures a major component
in overall simplification quality, underscoring \samsa's value. These promising results also motivate a future combination of SAMSA with classifier-based metrics.

\vspace{-0.1cm}
\section{Conclusion}
\vspace{-0.2cm}
We presented the first structure-aware metric for text simplification, \samsa, and the first evaluation experiments that directly target the structural simplification component, separately from the lexical component.
We argue that the structural and lexical dimensions of simplification are loosely related, and that TS evaluation protocols should assess both.
We empirically demonstrate that strong measures that assess lexical simplification quality (notably SARI), fail to correlate with human judgments when structural simplification is performed by the evaluated systems.
Our experiments show that \samsa\ correlates well with human judgments in such settings, which demonstrates its usefulness for evaluating and tuning statistical simplification systems, and shows that structural evaluation provides a complementary perspective on simplification quality.

\vspace{-0.1cm}
\section*{Acknowledgments}
\vspace{-0.1cm}
We would like to thank Zhemin Zhu and Sander Wubben for sharing their data, as well as the annotators for participating in our evaluation and UCCA annotation experiments.
We also thank Daniel Hershcovich and the anonymous reviewers for their helpful comments. This work was partially supported by the Intel Collaborative Research Institute for Computational Intelligence (ICRI-CI) and by the Israel Science Foundation
(grant No. 929/17), as well as by the HUJI Cyber Security Research
Center in conjunction with the Israel National Cyber
Bureau in the Prime Minister's Office.

\bibliography{biblionaaclhlt2018_source}
\bibliographystyle{acl_natbib}

\end{document}